# ESCELL: EMERGENT SYMBOLIC CELLULAR LANGUAGE

*Aritra Chowdhury\*, James R. Kubricht\*, Anup Sood\*\*, Peter Tu\*, Alberto Santamaria-Pang\**

\*Artificial Intelligence, GE Research, Niskayuna, USA
\*\*Biology and Applied Physics, GE Research, Niskayuna, USA

## ABSTRACT

We present ESCELL, a method for developing an emergent symbolic language of communication between multiple agents reasoning about cells. We show how agents are able to cooperate and communicate successfully in the form of symbols similar to human language to accomplish a task in the form of a referential game (Lewis' signaling game). In one form of the game, a sender and a receiver observe a set of cells from 5 different cell phenotypes. The sender is told one cell is a target and is allowed to send one symbol to the receiver from a fixed arbitrary vocabulary size. The receiver relies on the information in the symbol to identify the target cell. We train the sender and receiver networks to develop an innate emergent language between themselves to accomplish this task. We observe that the networks are able to successfully identify cells from 5 different phenotypes with an accuracy of 93.2%. We also introduce a new form of the signaling game where the sender is shown one image instead of all the images that the receiver sees. The networks successfully develop an emergent language to get an identification accuracy of 77.8%.

## 1. INTRODUCTION

Mainstream deep learning approaches are hard to interpret. This is because deep learning relies on feature representations in continuous high dimensional spaces that are difficult for humans to comprehend. Human language is the communication channel through which people understand and cooperate with each other. The protocols for communication have been developed for thousands of years and each population has given rise to their own set of languages that have emerged out of the necessity for social collaboration among human agents. In this work, we introduce the idea of emergent languages between artificial agents to collaborate on understanding the nature of cell biology. In particular, we work with a dataset consisting of cells stained with 4 different markers – CD3, CD20, CD68 and Claudin1. In addition, we have cells that do not stain for any of the 4 markers. Therefore, in total we have a set of 5 concepts that are categorized by their phenotypical characteristics. A coherent language in the form of symbols is observed to be emergent from a referential Lewis' signaling game [1]. Understanding the language of cells in this manner, will help us understand biology in a symbolic manner and it can be used as another vehicle for scientific discovery.

The method is based on research involving multi-agent coordination communication games. The agents in such games start as *tabula rasa*, but through the constraints of the game, they can infer knowledge about the game world leading to the emergence of an artificial symbolic language. The symbols generated from our emergent language framework (ESCELL) show that the agents are able to collaborate on the referential game. This is because, the agents associate different cells to different phenotypes.

Deep learning approaches for classification or segmentation do not provide implicit methods for probing and understanding how they make predictions and decisions. This makes it difficult for them to be adopted reliably for making important decisions in healthcare. Our approach of using emergent languages can lead the way to make neural networks more transparent and help medical practitioners trust artificial systems to aid them in making conclusions on diagnosis and prognosis of diseases. Moreover, this work can be extended to ground the emergent language in natural human language for further interpretability of deep learning models.

## 2. PREVIOUS WORK

This framework is inspired by Lazaridou et. al [2], where they introduce the idea of using referential games for multi-agent cooperation and show emergence of artificial language. They also discuss ideas to ground the symbols in natural languages. Havrylov et al. [3] extend these ideas to incorporate a sequence of symbols to further approximate sentence formation in emergent languages. The sequence of symbols is modelled using a type of recurrent neural network called LSTMs. They also consider introducing elements of natural language priors in the models using captions. Cogswell et al. [4] introduce compositional generality in the emergent languages among multiple agents.

Larazidou et al. [5] present a series of studies investigating the properties of the protocols from the language generated by the agents, who are exposed to symbolic and image data. In this work, we make an attempt to formulate and extend the ideas of emergent language communication to come up with a method for generating an emergent language of cells.

## 3. DATA

A hyperplex immunofluorescence microscopy platform (Cell DIVE™) is used in this work. It allows subcellular imaging of over 60 markers in a single 5μm formalin fixed paraffin embedded (FFPE) tissue section [6]. It involves multiple sessions of staining, imaging and signal inactivation, illumination correction, registration and autofluorescence removal.

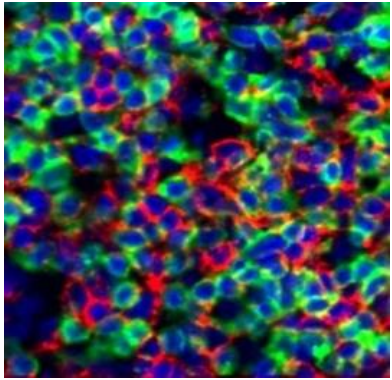

Fig. 1: Color images for immune cell markers for CD20 (red), CD3 (green), and Nuclei DAPI (blue).

We use a colon cohort in this analysis whose collection methods and details are provided in [6]. In this work, we use 4 cell markers – CD3, CD20, CD68 and Claudin1. In addition, we also use cells that don't stain positive for any of the 4 markers as a control group. Seven statistical intensity and shape based features are extracted from each cell marker encoding the information relevant to each marker. This makes it a total of 28 features – 1 set of 7 features for each of the 4 markers.

## 4. METHODS

ESCELL consists of a single symbol communication game where two agents play the Lewis' signaling referential game. The game is structured as follows.

1. There is a set of vectors representing the cells $\{v_1, v_2, ..., v_N\}$. $K$ vectors $\{u_1, u_2, ..., u_K\}$ are drawn at random from each of the $K$ different concepts. One of them $u_t$ is chosen to be a target $t \in \{1, 2, ..., K\}$
2. The sender network sees the set of $K$ sampled images and generates a symbol from a vocabulary of size $V$.
3. The receiver network, oblivious of the target $u_t$, sees the sender's symbol and a random permutation of the $K$ sampled images and tries to guess the target image.
4. Both the sender and receiver networks are rewarded for the correct guess, and penalized in case of a wrong answer.

This framework is inspired by Lazaridou et al. [2]. This is denoted as Experiment 1 in the results shown in Section 5. In addition to this setup, we also perform experiments with the setup used by Havrylov et al. [3]. Here, instead of showing all the $K-1$ distractor images to both the sender and receiver, the sender only sees the target image. This is denoted as Experiment 2, which is more challenging and realistic than Experiment 1. The symbol generated by the sender in step 2 of the framework requires sampling over the vocabulary. Sampling is not a continuous function and therefore, gradient computation and backpropagation are not possible. Using reinforcement learning is a possibility. However, training becomes much harder in that case. Instead, Gumbel softmax [7] estimators are used in place of sampling to allow for end-to-end differentiation. The methodology is shown in the following figure.

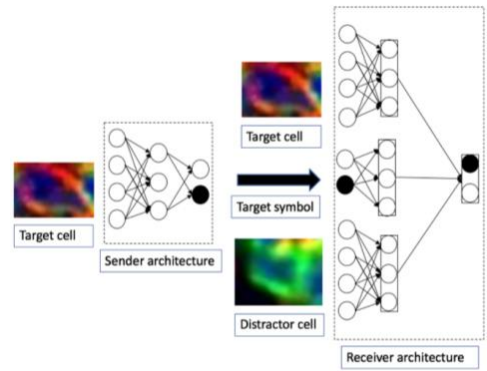

Fig. 2: ESCELL framework. A target cell is shown to the sender which is encoded as a neural network. The sender produces a symbol to represent the cell. This symbol is sent to the receiver. The receiver sees the target cell and a distractor cell and is asked to pick out the target based on the symbol sent by the sender. The entire sender and receiver architectures are trained in an end to end manner.

Fig. 2 shows a toy example of how ESCELL works. It actually shows Experiment 2, where the sender only sees the target image and has no information of the distractor image. The goal of this example is for the receiver to correctly identify the target cell based on the symbol that the sender transmits to the receiver. The target cell representation is forwarded through a sender neural network architecture. The sender outputs a symbol using Gumbel softmax relaxation. This symbol is fed to the receiver. The receiver also consists of a neural network that takes as input the target image, the distractor image and the symbol. It combines these inputs using a receiver neural network architecture and guesses the target.

We use negative log likelihood loss (similar to classification) to compute the loss of the receiver. This loss is backpropagated through the receiver and sender architectures to train the ESCELL framework. The next section describes the experimental settings and results using this framework.

## 5. EXPERIMENTS AND RESULTS

The dataset used in the experiments is obtained from the data described in Section 2. We extract 28 features from the cells – 7 quantitative shape and intensity based statistics from each immune marker. There are a total of 4125 samples (138 from CD3, 132 from CD20, 177 from CD68, 391 from Claudin1 and 3287 from Negative). We divided the data into a training, validation and testing with stratified splits of 64%, 16% and 20% respectively. The Emergence of language in games (EGG) toolkit [7] was used and modified to implement the referential games in the following experiments:

1. The traditional referential game described in the framework in Section 4. Here the sender and receiver both observe the $K$ images sampled independently from each concept. The receiver observes a permutation of the sampled images shown to the sender and must guess which of the $K$ images is the target $u_t$.

2. A modification of the sender receiver referential game inspired by [3], where the sender is only shown the target image $u_t$, generates a symbol $s$ from the vocabulary. Similar to the first experiment, the receiver must now choose the correct image from the samples using $u_t$.

The sender and receiver are encoded as feedforward neural networks in these experiments. The architecture embeds the input vector in a "game-specific" embedding space of size 15, followed by 1-D convolutional layer with sigmoid non-linearity. The resulting feature maps are sent through another non-linear filter to produce scores over a vocabulary of size 100. The activation of this layer is encapsulated by a Gumbel Softmax relaxation to produce a single symbol. The symbol is a one-hot vector over the vocabulary space.

The receiver architecture takes the symbol, the target and 4 distractor images sampled from different concepts in random order. The symbol and the images are embedded in a "game-specific" embedding space of size 15. A dot-product is computed between the symbol embedding and the image embeddings. The dot products are then converted to log probabilities using a log softmax layer across the 5 outputs of the cross product. This output points to the target cell image. The loss is computed as the negative log likelihood using this output and the target one-hot vector. Fig. 2 shows a visualization of the framework which is inspired from [2] and [3].

Table 1 shows the results of the two experiments in terms of their identification accuracy, fraction of symbols used in the language, and dominant symbols used for each of the markers with their corresponding percentage within the same marker. We observe that the accuracy is higher in the case of experiment 1 (5 sender images) than in experiment 2 (1 sender image). This is intuitive because the sender has access to all distractor images and can adequately discriminate between them. In addition, we see that experiment 1 uses approximately 33% of the symbols compared to experiment 2 and is far more efficient. This can be attributed to the same reason.

Table 1: Results of signaling game on the two experiments. In Experiment 1, both the sender and receiver observe 5 input cells, whereas in Experiment 2, the sender observes only the target cell and the receiver observes the 5 input cells from which to pick the target.

|  | Experiment 1 | Experiment 2 |
|---|---|---|
| Identification accuracy (%) | 93.18 | 78.78 |
| Fraction of symbols used (%) | 5 | 15 |
| CD3 majority symbol (% fraction) | 62 (96.43) | 32 (100) |
| CD20 majority symbol (% fraction) | 50 (85.18) | 76 (96.29) |
| CD68 majority symbol (% fraction) | 55 (94.44) | 81 (86.11) |
| Claudin1 majority symbol (% fraction) | 84 (82.50) | 89 (33.75) |
| Negative majority symbol (% fraction) | 97 (61.67) | 32 (43.64) |

We also note that both the experiments associate different symbols with each of the protein markers, and the corresponding percentages of majority symbols are quite high with the exception of Claudin1 and Negative (denoted as *None* in the experiments) in experiment 2. We also observe that symbol 32 appears as majority for both CD3 and Negative because few cells have weak CD3 staining.

Fig. 3 shows a distribution of symbols with respect to each of the protein markers in the form of violin plots.

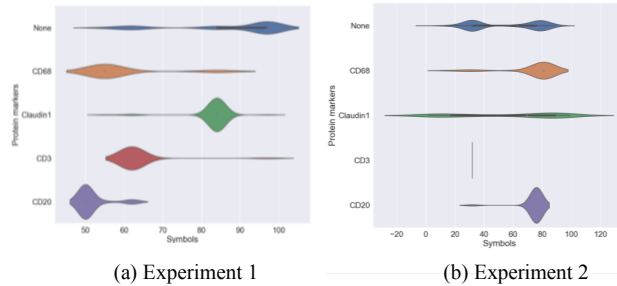

(a) Experiment 1      (b) Experiment 2

Fig. 3: Violin plots of the distributions of symbols from 1-100 with respect to the protein markers in (a) Experiment 1 and (b) Experiment 2

As we can see from the plots, each protein marker seems to be associated with a distinct symbol. The symbols at which we see the violin plots concentrate correspond to the majority symbols observed in Table 1.

The results of Experiment 1 are better than Experiment 2 which is reflected by the identification accuracies in Table 1 and the clustering of the symbols in Fig. 3. This is expected because in Experiment 1, the sender has access to the distractor cells and can adequately differentiate between the target and distractors. In Experiment 2, the sender only sees the target cell and this setting is therefore much harder. We also observe that the symbols for Negative (represented as *None* in the plots) are more distributed than the other protein markers. This could be due to the fact that the cells that don't stain positive for any of the markers in this work, could be positive for other markers. This could also be due to non-specific staining in the Negative class. It would be interesting to see if the symbols generated in the Negative class could correspond to any new phenotypes.

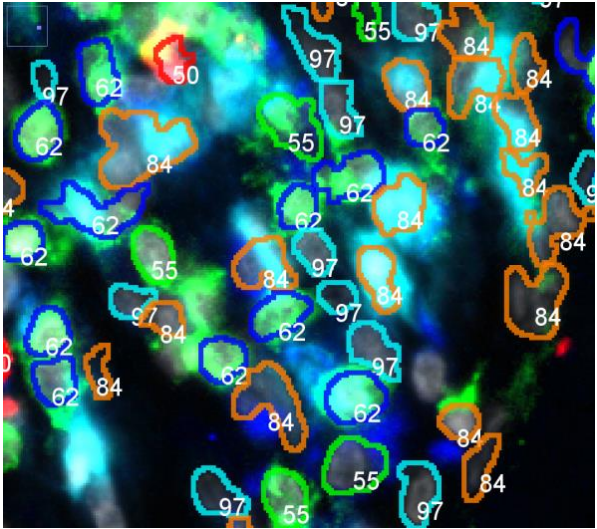

Fig. 4: Color composite image: CD20 (B-Cell), CD3 (T-Cell), CD68 (Macrophage), Claudin1 and DAPI (Cell Nuclei), in Red, Green, Blue, Cyan and Gray respectively. The segmented stroma nuclei is overlaid with the symbol identifier for each cell. We observe that the different cell phenotype markers are associated with distinct symbols, but are same for the cells with the same marker. The symbols generated correspond to Experiment 1 in Table 1 and Fig. 3(a).

Fig. 4 shows a visualization of a color composite image with the markers along with the nuclei. The markers are represented as different colors – CD20 (Red), CD3 (Green), CD68 (Blue) and Claudin1 (Cyan). The cell nuclei (DAPI) is represented as gray. We observe that ESCELL is able to successfully associate symbols with different immune cells with a high degree of accuracy. The overlaid symbols are generated using the model in Experiment 1 where the sender architecture observes all the distractor images. Even though the cells within the same marker have different shape and intensity characteristics, the ESCELL framework is able to develop a coherent symbolic language to communicate the cell phenotypes adequately.

The dominant markers in Table 1 show up consistently in a section of the tissue sample in Fig. 4. The results in Table 1, Fig. 3 and Fig. 4 show that ESCELL can be used a symbolic language framework for cell biology.

## 6. CONCLUSION

We introduce an emergent symbolic cellular language framework known as ESCELL. It is formulated as a 2-agent game in which agents communicate with each other using the language of symbols. The framework is based on a referential signaling game where the task is to identify the target cell among distractor cells. The framework consists of a sender and receiver. Both are formulated as end to end deep neural networks. We show that the symbols generated by the sender are distinct for each cell phenotype. This is one step towards making deep learning methods more interpretable by introducing symbols as a form of communication language between artificial agents reasoning about the biology of cells, in particular the expressed phenotypical characteristics.

There is scope for a lot of potential work going forward. Convolutional neural networks maybe trained to generate symbols in an end to end manner to generate symbols instead of using feature extraction methods. One very important step towards true interpretability would be to ground the symbols with natural human language. In addition, the game could be extended to include other learning settings such as supervised (eg. classification) and unsupervised learning (e.g. segmentation).